\documentclass[letterpaper,10pt,conference]{ieeeconf}

\IEEEoverridecommandlockouts
\usepackage{graphicx}
\usepackage{booktabs}
\usepackage{amsmath,amssymb}
\usepackage[hyphens]{url}
\usepackage{cite}
\usepackage{placeins}

\newcommand{\obs}{o}
\newcommand{\lang}{\ell}
\newcommand{\wm}{\mathcal{W}_{\theta}}
\newcommand{\goal}{I^{\star}}
\newcommand{\motion}{\mu}

\title{\LARGE \bf
WAVE-Go: World-Model Navigation with Adaptive Execution for Wheel-Legged Robots
  
}

\author{Mingyi Li$^{1}$, Ji Li$^{2,\dagger}$, Zhihao Ouyang$^{3,\dagger}$, Yage He$^{4}$, B\"orje F. Karlsson$^{5}$\\[1mm]
\small $^{1}$Beijing Institute of Technology \quad $^{2}$The University of Hong Kong\\
\small $^{3}$GAC R\&D Center\\
\small $^{4}$LimX Dynamics \quad $^{5}$Beijing Academy of Artificial Intelligence (BAAI)\\
\small $^{\dagger}$Equal contribution}

\begin{document}
\maketitle
\thispagestyle{empty}
\pagestyle{empty}

\begin{abstract}

World models can anticipate the consequences of navigation actions, but predicted action sequences may become invalid during execution, especially when wheel-legged robots encounter dynamic obstacles or change locomotion modes. We propose WAVE-Go, an image-goal navigation framework that separates world--action prediction from interruptible command execution. Its executor adaptively selects an action prefix and cancels pending commands when updated observations invalidate execution. A conditional-risk formulation specifies prefix selection under an estimated cumulative failure budget, while posture and locomotion-mode transitions require clearance, stability, and task-evidence checks. In the reported navigation evaluation, WAVE-Go achieves 74.1\% in-distribution success and 63.3\% dynamic out-of-distribution success, exceeding the strongest baseline by 4.7 and 7.7 percentage points, respectively, while reducing collisions from 4.4 to 2.9 per 100\,m. Compared with interruptible fixed four-command execution, WAVE-Go raises success by 4.0 percentage points while reducing replanning frequency by 51.2\% and collision rate by 6.5\%. Execution ablations also show that runtime interruption improves success, collision rate, and reaction latency at the cost of additional replanning. These results support adaptive, interruptible execution as a means of balancing navigation performance and planning overhead. Code is available at \url{https://github.com/vigorlee/wave-go}.

\end{abstract}


\section{Introduction}

World models support visual navigation by predicting the consequences of robot actions \cite{nwm,dinowm,li2026himmhumaninspiredlongtermmemory}. Joint world--action models pair candidate commands with anticipated successor states \cite{futurenav,navwam,wamnav}. For image-goal navigation, these predictions connect the current observation to a desired visual destination. Their value, however, depends on whether the proposed commands remain appropriate as the robot moves and receives new observations. Prediction and execution therefore require distinct decisions: proposing a plausible future does not determine how long the robot should follow it.

This distinction matters for wheel-legged robots, whose executable motions depend on both scene geometry and locomotion mode. A sequence suitable for an open corridor may become invalid near clutter, after target occlusion, or during a posture transition. Long action chunks reduce planning frequency but prolong commitment to a prediction; short chunks permit frequent revision at additional inference cost. Mode changes add a discrete dependency: subsequent driving must await confirmation of the requested configuration. The central problem is to choose an execution length that balances this commitment against changing conditions, while retaining the ability to interrupt pending commands.

Predictive uncertainty alone does not determine how long a robot should follow an action sequence. A confidently predicted motion may still approach an obstacle or violate a locomotion-mode constraint, whereas uncertain visual details may have little effect on a short motion through verified free space. Uncertainty-adaptive chunking adjusts execution length using prediction-confidence cues \cite{adaptivechunk}, but selecting an execution prefix also requires assessing the consequences of continuing under current conditions. WAVE-Go addresses this need by combining predictive cues with execution conditions to estimate failure risk over successive commands. Constraining cumulative risk over the prefix allows execution length to reflect both the likelihood of failure and the exposure introduced by continued execution.

To address these challenges, we propose WAVE-Go, a navigation framework that couples joint world--action prediction with adaptive, interruptible execution. Conditioned on the goal image, recent observations and robot motion, and a structured semantic task state, the predictor generates candidate actions and their associated future-state features. The executor selects the longest valid prefix from 4, 8, and 16 commands under an estimated cumulative failure budget. This selection sets an execution limit rather than a commitment to complete the prefix: updated RGB-D/LiDAR observations can trigger cancellation of pending commands, while target and mode checks enforce task prerequisites. The executor requests a new prediction when a prefix ends or is interrupted; if execution involves a posture or locomotion-mode transition, it first confirms the resulting configuration so that subsequent actions reflect the robot's updated state. By combining risk-based prefix selection with observation-based interruption, WAVE-Go adapts how long it follows a prediction while maintaining feedback throughout execution.

Our contributions are:
\begin{itemize}
\item \textbf{Cumulative-risk prefix authorization:} An execution formulation that combines conditional failure estimates and validity checks to determine how much of a predicted action sequence to execute before replanning.
\item \textbf{Interruptible, mode-aware execution:} A command interface that revalidates pending actions and treats posture and locomotion-mode transitions as execution boundaries requiring configuration confirmation.
\item \textbf{Performance--cost trade-offs:} Results show improved navigation with fewer replanning calls than short fixed-prefix execution, while interruption improves command responsiveness at an additional replanning cost.
\end{itemize}

\begin{figure*}[t]
\centering
\includegraphics[width=0.78\textwidth]{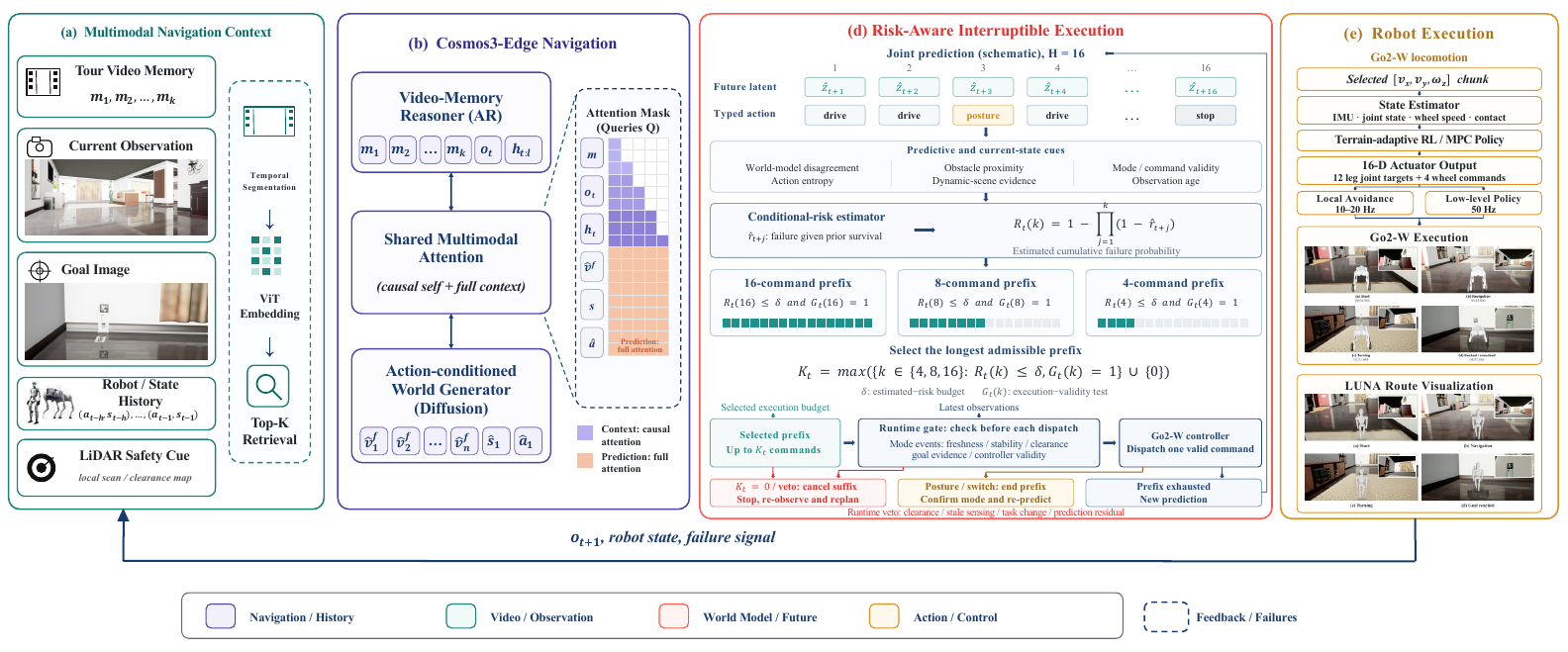}
\caption{WAVE-Go architecture: (a) multimodal context, (b) context fusion and joint world--action prediction, (c) predicted actions, successor latent states, modes, and auxiliary cues, (d) conditional-risk prefix selection and interruptible execution (Eqs.~\ref{eq:step-risk}--\ref{eq:adaptive-k}), and (e) the Go2-W control interface with Go2-W and LUNA execution illustrations. Runtime checks can cancel pending commands; posture or mode events require configuration confirmation and a new prediction. The rollout is schematic; images are simulator views.}
\label{fig:v3-overview}
\vspace{-8pt}
\end{figure*}

\section{Related Work}

\paragraph{World models for navigation}
World models support navigation by anticipating the outcomes of candidate actions. NWM predicts future visual observations to support trajectory optimization and candidate ranking \cite{nwm}, while DINO-WM predicts pretrained visual features and optimizes action sequences toward a goal representation \cite{dinowm}. Related work explores representations and rollout strategies that address prediction cost and temporal consistency \cite{raenwm,arforcing,mwm}. SparseVideoNav uses sparse future video generation to guide long-horizon navigation with reduced inference latency \cite{sparsevideonav2026}. ImagiNav decouples visual planning from actuation by generating future videos and extracting executable trajectories through inverse dynamics \cite{imaginav2026}. Beyond navigation, Dreamer demonstrates world-model-based learning across diverse control tasks \cite{hafner2025dreamer}. WAVE-Go addresses the execution interval following prediction, when new observations may invalidate previously selected commands.

\paragraph{Joint world--action prediction}
Joint world--action models couple future-state prediction with action generation. FutureNav combines complementary learning objectives for vision-and-language navigation \cite{futurenav}, and NavWAM jointly models navigation states and actions \cite{navwam}. WAM-Nav further distinguishes the horizons used for action generation and future-state prediction \cite{wamnav}. Related formulations extend to multimodal trajectory prediction in NavWM and aerial navigation in WorldVLN \cite{navwm2026,worldvln}. Fast-WAM distinguishes video co-training from explicit test-time imagination, showing that competitive manipulation performance can be retained while omitting future generation at inference \cite{fastwam2026}. These results motivate distinguishing the benefits of predictive representations from the cost and role of online prediction. Building on the joint prediction interface, WAVE-Go determines how much of a candidate action sequence to execute and when to interrupt it on a wheel-legged platform.

\paragraph{Action chunks and feedback control}
Goal-conditioned navigation policies such as NoMaD emphasize closed-loop operation \cite{nomad}, while NavDP studies navigation across robot embodiments \cite{navdp}. Action Chunking with Transformers generates action sequences to reduce sequential policy decisions \cite{act}, and adaptive chunking uses predictive uncertainty to adjust execution horizons \cite{adaptivechunk}. Real-time chunking overlaps action execution with inference and preserves committed actions to maintain continuity across chunk boundaries \cite{black2025rtc}. REMAC further addresses intra-chunk inconsistency between predicted actions and current observations through masked action chunking and learned corrections \cite{wang2026remac}. WAVE-Go combines risk-budgeted execution prefixes with validity checks before each command and early interruption when execution conditions change. For posture and locomotion-mode transitions, these checks additionally incorporate clearance, stability, and task readiness, followed by confirmation of the resulting configuration before subsequent motion.

\section{Task Formulation}

We formulate image-goal navigation as a partially observed control problem. At time \(t\), the observation \(\obs_t=(I_t,D_t,L_t)\) contains an RGB image, depth, and local range measurements. The remaining inputs are a goal image \(\goal\), optional language \(\lang\), proprioceptive state \(x_t\), and locomotion mode \(m_t\in\mathcal M\). The observation history cannot fully resolve occlusions or future scene changes, so execution uses updated observations rather than relying solely on a predicted trajectory.

The model uses a semantic task state \(c_t\) and a motion-context token \(\motion_t\). The former records goal progress and task dependencies; the latter encodes recent visual motion, proprioception, actions, and optional tracking features. We write \(q_{u:v}=(q_u,\ldots,q_v)\) for a time-indexed sequence and define
\begin{equation}
\begin{aligned}
\mathcal H_t^{\mathrm{mot}}
  &= (I_{t-h:t},x_{t-h:t},a_{t-h:t-1},\tau_{t-h:t}),\\
\motion_t
  &= E_{\mathrm{mot}}(\mathcal H_t^{\mathrm{mot}})\in\mathbb R^{64}.
\end{aligned}
\label{eq:motion-token}
\end{equation}
Here \(h\) is the number of preceding observation intervals and \(\tau\) denotes optional target-tracking features. The history contains \(h+1\) observations and the \(h\) actions between them; it excludes the action to be selected at time \(t\).

The prediction context and joint rollout are defined as
\begin{equation}
\begin{aligned}
\mathcal C_t
  &= (\obs_{t-h:t},\goal,\lang,c_t,\motion_t,m_t),\\
\hat y_{t,j}
  &= (\hat z_{t+j},\hat a_{t+j-1},\hat m_{t+j-1},\hat\nu_{t+j-1}),\\
(\hat y_{t,1},\ldots,\hat y_{t,H})
  &= \wm(\mathcal C_t).
\end{aligned}
\label{eq:world-action}
\end{equation}
For \(j=1,\ldots,H\), each predicted step pairs a candidate action \(\hat a_{t+j-1}\) and its mode \(\hat m_{t+j-1}\) with the successor latent state \(\hat z_{t+j}\). The auxiliary features \(\hat\nu_{t+j-1}\) provide uncertainty and command-validity cues. We specify \(H=16\). This joint-output interface does not imply conditioning on an externally prescribed action sequence. The executor selects at most \(K_t\leq H\) commands and cancels pending commands when updated observations invalidate execution.

\section{Method}

WAVE-Go connects joint world--action prediction to adaptive, interruptible execution. Its core execution mechanism uses a trained and calibrated conditional-risk model to select action prefixes, checks updated observations to interrupt invalid commands, and verifies posture and locomotion-mode transitions before subsequent motion. We first introduce predictor adaptation, multimodal context encoding, and joint learning objectives, which define the candidate actions and future-state features available to the executor. We then describe our proposed hybrid command interface, risk-based prefix selection, transition checks, and multi-rate execution.

\subsection{Predictor Adaptation}

The post-training formulation uses temporally aligned observation--action transitions:
\begin{equation}
\begin{aligned}
\mathcal{D}_i={}&
(\obs_{t-h:t},\goal,\lang,c_t,x_t,m_t,\\[-1mm]
&a_{t:t+H-1},m_{t:t+H-1},\obs_{t+1:t+H}).
\end{aligned}
\label{eq:data-tuple}
\end{equation}
Each tuple links an observation history to executed actions, locomotion modes, and the resulting observations. Goal images and terminal-event annotations distinguish reaching a visually similar location from completing the task. Failed transitions provide labels for consequence prediction and failure estimation when their causes are observable.

We first adapt the baselines' visual--language--action pairs to embodiment-specific transitions covering occlusions, dynamic obstacles, target reacquisition, posture changes, and locomotion-mode transitions. This adaptation targets the observation and action distribution of the wheel-legged platform. The available frozen-model trace does not measure the benefit of this curriculum.

Transition sampling is stratified around target discovery, occlusion, obstacle appearance, stopping, and mode changes to limit the dominance of forward-motion examples. Matched successful and failed transitions provide additional supervision where they are available.

\subsection{Multimodal Context}

The semantic reasoner combines the goal image, selected observations, optional language, the previous task state, and the available robot capabilities. Its output follows a fixed schema:
\begin{equation}
c_t=(p_t,s_t,u_t,\kappa_t,b_t),
\label{eq:semantic-state}
\end{equation}
where \(p_t,s_t,u_t,\kappa_t,b_t\) encode goal progress, accumulated evidence,
the next subgoal, confidence, and ordered dependencies, respectively. A Qwen-3-VL backbone produces this state without generating high-rate velocity commands \cite{qwen25vl}.

The goal, reasoner state, and motion token are fused by cross-attention:
\begin{align}
h_t &= E_{\mathrm{vis}}(\obs_{t-h:t}), \nonumber\\
\tilde h_t &= \mathrm{Attn}
\left(h_t,\,[E_{\mathrm{goal}}(\goal);
E_{\mathrm{lang}}(\lang,c_t);\motion_t]\right).
\label{eq:fusion}
\end{align}
The fused representation integrates current visual observations with the navigation goal and recent robot motion, using a learned null token when language input is unavailable. Auxiliary objectives supervise proprioceptive state, short-range egomotion, and target displacement, while promoting consistency between actions and state transitions. The world--action head is conditioned on the structured task state, which encodes task-relevant information from semantic reasoning in a fixed schema.

The proposed encoder applies stop-gradient operations to limit the encoding of goal identity in the motion token. Probes for motion and locomotion mode, together with a control that shuffles motion tokens, assess whether the representation captures motion information aligned with the observation history.

\subsection{World--Action Learning}

The predictive objective combines future-state modeling, inverse dynamics,
action generation, and mode prediction:
\begin{equation}
\begin{aligned}
\mathcal{L}={}&
\lambda_f\mathcal{L}_{\mathrm{future}}+
\lambda_i\mathcal{L}_{\mathrm{inverse}}+
\lambda_a\mathcal{L}_{\mathrm{action}}\\
&+\lambda_m\mathcal{L}_{\mathrm{mode}}+
\lambda_c\mathcal{L}_{\mathrm{consistency}}.
\end{aligned}
\label{eq:loss}
\end{equation}
Here, $\mathcal{L}_{\mathrm{future}}$ supervises future latent-state prediction, $\mathcal{L}_{\mathrm{inverse}}$ supervises inverse dynamics, and $\mathcal{L}_{\mathrm{action}}$ supervises action generation. The mode loss $\mathcal{L}_{\mathrm{mode}}$ supervises locomotion-mode prediction, while $\mathcal{L}_{\mathrm{consistency}}$ encourages consistency between predicted actions and their associated future states. This objective builds on joint world--action prediction formulations \cite{futurenav,navwam}.

\subsection{Hybrid Commands}

The action interface integrates continuous drive and posture commands with discrete locomotion-mode events. Controller-specific validity masks enforce the command constraints associated with each mode. Posture and mode-transition events mark the end of the current execution prefix, as subsequent drive commands must be conditioned on the resulting robot configuration.

\subsection{Conditional-Risk Prefix Selection}\label{sec:risk}

Longer action prefixes reduce planning frequency but extend execution without a new world--action prediction \cite{adaptivechunk}. To balance planning frequency against the risk of continued execution, we estimate cumulative failure risk over each candidate prefix and use it to determine the execution length. Let \(F_{t+j}\) denote a failure event following command \(a_{t+j-1}\), including collision, sustained lack of goal progress, or an invalid mode transition. These events are defined using thresholds fixed before calibration. A trained failure model estimates the conditional probability of failure at offset \(j\), given that all preceding commands have completed without failure:
\begin{equation}
\begin{aligned}
\hat r_{t+j}={}&
\Pr\!\left(F_{t+j}=1 \,\middle|\,
\tilde h_t,\hat a_{t:t+H-1},\right.\\[-1mm]
&\hspace{23mm}\left.\hat\nu_{t:t+H-1},
F_{t+1:t+j-1}=0\right).
\end{aligned}
\label{eq:step-risk}
\end{equation}
The model combines world-model disagreement, action entropy, obstacle proximity, dynamic-scene evidence, mode status, and sensor age. The conditional survival probabilities yield the cumulative failure estimate for a length-\(k\) prefix:
\begin{equation}
R_t(k)=1-\prod_{j=1}^{k}(1-\hat r_{t+j}).
\label{eq:prefix-risk}
\end{equation}
This formulation accounts for survival through preceding commands without assuming independent failures. Given the candidate lengths \(\mathcal K=\{4,8,16\}\), the executor selects the longest prefix that satisfies the estimated risk budget \(\delta\) and the validity test \(G_t(k)\):
\begin{equation}
\begin{aligned}
K_t=\max\big(&\{k\in\mathcal{K}:R_t(k)\leq\delta,\\[-1mm]
&G_t(k)=1\}\cup\{0\}\big).
\end{aligned}
\label{eq:adaptive-k}
\end{equation}
Here, \(G_t(k)\) checks command validity, mode preconditions, and observation freshness. If no candidate is admissible, \(K_t=0\), and the executor requests a stop and renewed sensing.

The selected \(K_t\) limits the number of command dispatches, while runtime checks can terminate execution earlier. Before each dispatch, the executor checks the latest observations for insufficient clearance, stale sensing, changed task state, or excessive prediction residuals; any violation cancels the remaining commands and triggers replanning. Posture and mode events also terminate the prefix, with subsequent driving requiring confirmation of the resulting configuration and a new prediction. Figure~\ref{fig:v3-overview}(d) summarizes this selection and interruption process.

The selector uses a trained and calibrated conditional-risk model. The calibration protocol reserves a separate scene split and evaluates cumulative risk at each candidate length, since stepwise calibration alone does not ensure prefix-level calibration. Reliability diagrams and expected calibration error can quantify agreement between predicted and observed failure frequencies \cite{guo2017calibration}. The resulting risk estimates depend on the learned conditional probabilities; \(\delta\) therefore specifies an estimated failure budget rather than a certified physical collision bound.

\subsection{Evidence-Gated Hybrid Transitions}

Posture and locomotion-mode transitions alter the conditions under which subsequent commands can be executed. A transition graph defines admissible mode changes and their prerequisites. For example, crouching requires the robot to be stationary and stable, with sufficient clearance from nearby obstacles. Task-specific transitions additionally depend on current goal evidence and satisfaction of prior task dependencies. These requirements are combined in a binary transition gate:
\begin{equation}
\begin{aligned}
G_t^{\mathrm{mode}}(m')={}&
g_t^{\mathrm{fresh}}g_t^{\mathrm{stable}}g_t^{\mathrm{clear}}\\[-1mm]
&\times g_t^{\mathrm{goal}}g_t^{\mathrm{valid}}(m').
\end{aligned}
\label{eq:transition-gate}
\end{equation}
For a candidate mode \(m'\), the five factors check observation freshness, stability, clearance, goal evidence, and controller validity, respectively. The gate contributes to the prefix-validity test \(G_t(k)\) in Eq.~\ref{eq:adaptive-k}, allowing the executor to verify each proposed transition independently before dispatch. Proposals rejected by the gate are recorded separately from transitions that fail after dispatch, so successful interventions are not counted as execution failures.

The executor handles invalid commands according to their type: rejecting a drive command triggers renewed sensing, while rejecting a posture or mode token does not select a substitute mode. If the semantic state is malformed, the executor retains the last validated state and suspends task progression.

For each command dispatched at offset \(j\), an execution record links the observed outcome to the corresponding prediction, risk estimates, and runtime decisions:
\begin{equation}
\begin{aligned}
\mathcal{E}_{t,j}=\big(&\obs_{t+j},\hat z_{t+j},
\hat a_{t+j-1},a^{\mathrm{exec}}_{t+j-1},\\[-1mm]
&\hat r_{t+j},R_t(j),\alpha_{t,j-1},
\gamma_{t,j-1},v_{t,j-1}\big).
\end{aligned}
\label{eq:audit-record}
\end{equation}
Here, \(\alpha\), \(\gamma\), and \(v\) denote authorization, gate outcomes, and the veto code, respectively. Rejected commands are logged separately because no observed successor state can be attributed to their execution. These records distinguish prediction errors, runtime interventions, and command-tracking failures.

\begin{table*}[!t]
\centering
\caption{Closed-loop navigation results with 99 rollouts per method per condition (ID, scene-OOD, and dynamic-OOD). Best and second-best values are \textbf{bold} and \underline{underlined}, respectively.}
\label{tab:main-results}
\footnotesize
\setlength{\tabcolsep}{3pt}
\begin{tabular*}{\textwidth}{@{\extracolsep{\fill}}lccccc@{}}
\toprule
Method & ID SR (\%)\(\uparrow\) & ID SPL (\%)\(\uparrow\) & Scene-OOD SR (\%)\(\uparrow\) & Dynamic-OOD SR (\%)\(\uparrow\) & Coll./100\,m\(\downarrow\)\\
\midrule
Action-only control & \(61.8\) & \(55.6\) & \(49.2\) & \(41.8\) & \(6.8\)\\
NWM + CEM~\cite{nwm} & \(64.7\) & \(57.9\) & \(55.8\) & \(50.2\) & \(5.1\)\\
FutureNav-style auxiliary WM~\cite{futurenav} & \(66.8\) & \(60.6\) & \(54.3\) & \(47.5\) & \(5.6\)\\
NavWAM-style joint W--A~\cite{navwam} & \(\underline{69.4}\) & \(\underline{63.0}\) & \(\underline{59.7}\) & \(\underline{55.6}\) & \(\underline{4.4}\)\\
WAVE-Go & \(\mathbf{74.1}\) & \(\mathbf{68.0}\) & \(\mathbf{64.8}\) & \(\mathbf{63.3}\) & \(\mathbf{2.9}\)\\
\bottomrule
\end{tabular*}

\vspace{-8pt}
\end{table*}

\subsection{Multi-Rate Execution}

To reduce repeated inference while preserving timely feedback, the runtime assigns distinct update schedules to command tracking, world--action prediction, and semantic reasoning. Command tracking and validity checks run at the controller frequency, world--action inference is triggered when a prefix expires or is cancelled, and semantic reasoning updates when task evidence changes. Before reuse, a buffered prediction is checked against observation freshness, the current mode, and the task state.

The key insight is that the frequency of feedback need not be tied to the frequency of prediction. A longer execution prefix can reduce model invocations while retaining checks before every command, allowing new observations to invalidate pending actions without waiting for the prefix to expire. Mode transitions provide an additional trigger for renewed prediction because they change the conditions governing subsequent motion. This scheduling strategy addresses the trade-off identified in the introduction by separating the decision to continue execution from the decision to generate a new action chunk. 

TensorRT acceleration and reasoning-model quantization are deployment options whose effects require separate measurement. Runtime evaluation at batch size one should distinguish model inference latency from observation-to-command latency and assess memory use and numerical agreement with the reference implementation. The reported action-selection rate includes task-level processing and therefore does not measure controller frequency or model throughput.

\section{Evaluation}

The evaluation addresses two questions: \textbf{(1)} Does WAVE-Go improve closed-loop navigation across different scene conditions? \textbf{(2)} How do execution-prefix selection and runtime interruption affect the trade-off between task performance and replanning frequency? We first introduce the experimental setup and evaluation metrics, then present quantitative results (Table~\ref{tab:main-results}) and execution-policy ablations (Table~\ref{tab:ablations}). Finally, we present qualitative results and additional measurements of action selection and stopping behavior.

\begin{table*}[!t]
\centering
\caption{Execution-policy ablation results for navigation success, collisions, command variation, replanning, and reaction latency. Best and second-best values are shown in \textbf{bold} and \underline{underlined}, respectively.}
\label{tab:ablations}
\footnotesize
\setlength{\tabcolsep}{3pt}
\begin{tabular*}{\textwidth}{@{\extracolsep{\fill}}lcccccc@{}}
\toprule
Execution rule & Interrupt & SR (\%)\(\uparrow\) & Coll./100\,m\(\downarrow\) & Cmd. variation\(\downarrow\) & Replans/100\,m\(\downarrow\) & Reaction p95 (s)\(\downarrow\)\\
\midrule
Fixed \(K=4\) & Yes & \(70.1\) & \(\underline{3.1}\) & \(0.41\) & \(48.6\) & \(\mathbf{0.18}\)\\
Fixed \(K=8\) & Yes & \(69.4\) & \(4.4\) & \(0.34\) & \(29.4\) & \(0.31\)\\
Fixed \(K=16\) & Yes & \(66.0\) & \(6.7\) & \(\mathbf{0.29}\) & \(\mathbf{17.8}\) & \(0.58\)\\
Action entropy only & Yes & \(70.8\) & \(3.8\) & \(0.33\) & \(27.0\) & \(0.29\)\\
World uncertainty only & Yes & \(\underline{71.9}\) & \(3.4\) & \(0.32\) & \(25.2\) & \(0.27\)\\
Calibrated risk & No & \(70.5\) & \(5.3\) & \(\underline{0.30}\) & \(\underline{20.1}\) & \(0.54\)\\
Full WAVE-Go & Yes & \(\mathbf{74.1}\) & \(\mathbf{2.9}\) & \(0.31\) & \(23.7\) & \(\underline{0.21}\)\\
\bottomrule
\end{tabular*}

\end{table*}

\subsection{Experimental Setup}

\paragraph{Conditions and comparisons}
The navigation comparison distinguishes in-distribution (ID), scene-OOD, and dynamic-OOD conditions, covering familiar scene distributions, unseen scenes, and changes in dynamic obstacles. Baselines comprise an action-only control, NWM with cross-entropy method planning \cite{nwm}, a FutureNav-style auxiliary world-model variant \cite{futurenav}, and a NavWAM-style joint world--action variant \cite{navwam}.

The comparisons use scene-disjoint data splits and paired episodes with identical initial poses, goals, obstacle schedules, sensor access, command limits, and timeouts. All execution-policy variants share the same frozen predictor. To isolate the effect of runtime interruption, full WAVE-Go and the no-interruption variant use the same trained and calibrated conditional-risk model, with cancellation of pending commands disabled only in the latter.

\begin{figure*}[!t]
\centering
\includegraphics[width=0.70\textwidth]{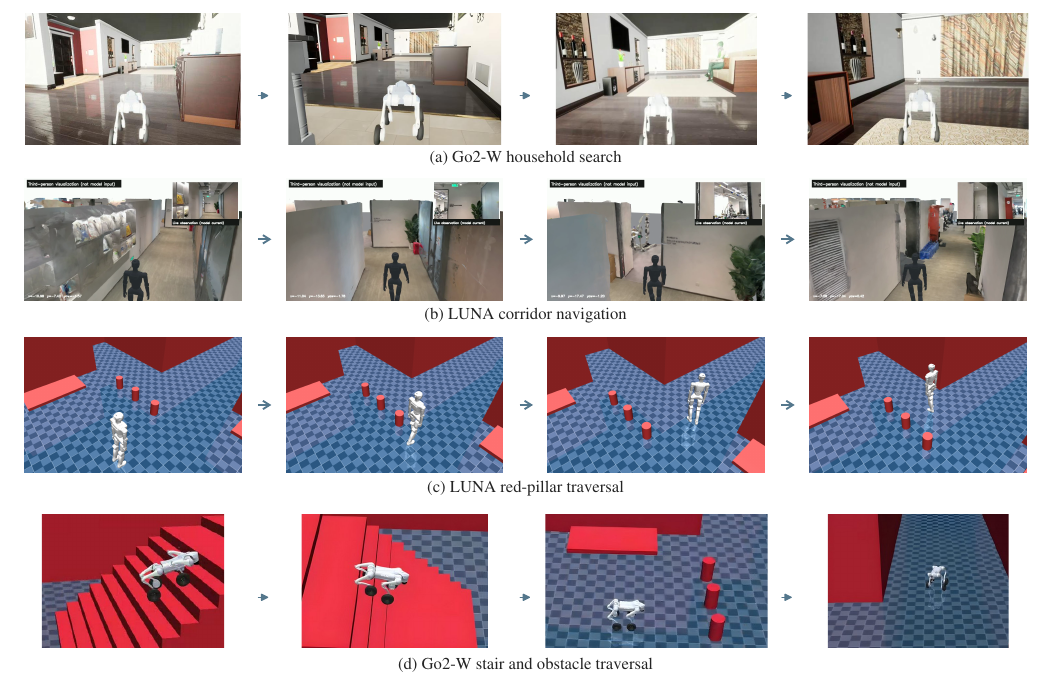}
\caption{Qualitative simulation sequences: (a) Go2-W household search, (b) LUNA corridor navigation, (c) waypoint-controlled LUNA red-pillar traversal in MuJoCo, and (d) Go2-W stair and obstacle traversal. Frames are ordered from left to right; third-person views in panels (b) and (c) are used only for visualization, not as model inputs.}
\label{fig:sim-exploration}
\vspace{-8pt}
\end{figure*}

\paragraph{Navigation metrics}
An episode succeeds if it reaches the specified goal region within the timeout and satisfies any terminal-mode requirement. An internal completion prediction alone does not establish success. For $N$ episodes with binary outcomes $S_i$, success rate is $\mathrm{SR}=N^{-1}\sum_{i=1}^{N}S_i$. Path efficiency is evaluated using success weighted by path length \cite{anderson2018evaluation}:
\begin{equation}
\mathrm{SPL}=\frac{1}{N}\sum_{i=1}^{N}S_i
\frac{l_i^{\star}}{\max(l_i^{\star},l_i)},
\label{eq:spl}
\end{equation}
where $l_i^{\star}>0$ is the shortest feasible reference path and $l_i$ is the executed path. Reference paths must respect the robot's traversability constraints. Failed episodes contribute zero to both SR and SPL, while successful detours reduce SPL. When a valid reference path is unavailable, SPL should be omitted; completion time can instead characterize delays caused by dynamic obstacles. Tables report SR and SPL as percentages.

\paragraph{Collisions and mode transitions}
The distance-normalized collision rate is $100\sum_i C_i/\sum_i l_i$, where $C_i$ counts unintended contacts in episode $i$ and distances are measured in meters. The sums include failed episodes. Each continuous contact counts once until separation, excluding intended ground and docking contacts. Interpreting this rate alongside success helps distinguish improved navigation from reduced exposure caused by early stopping. For mode transitions, a proposal rejected before dispatch is distinct from a dispatched transition that fails its preconditions. The latter rate uses all dispatched transitions as its denominator; separate rejection counts describe gate activity. These transition-level outcomes are not quantified in the current tables. Docking and sustained electrical charging require separate terminal-state evidence.

\paragraph{Execution metrics}
Replanning frequency is the number of model invocations per 100\,m traveled, including calls that produce discarded speculative predictions. This metric quantifies how often planning occurs, not the total time spent on inference. Reaction latency is measured from the availability of an observation that invalidates pending commands to the publication of a stop or replacement command. We report its 95th percentile to characterize delayed command responses; physical stopping time additionally depends on robot dynamics. In the no-interruption condition, pending-command cancellation is disabled, while the independent emergency stop remains active.

Command variation measures the mean second difference of normalized drive commands at a common interval $\Delta t$:
\begin{equation}
V_u=\frac{1}{M-2}\sum_{j=2}^{M-1}
\|\bar u_{j+1}-2\bar u_j+\bar u_{j-1}\|_2,
\label{eq:command-variation}
\end{equation}
for a sequence of $M\geq3$ commands. Each drive channel is normalized by its common command limit, and differences are evaluated only within the same episode. This dimensionless measure captures variation in commanded motion; estimating physical jerk would additionally require measured motion and time scaling. Success, collisions, reaction latency, and replanning must be considered together, since smoother commands or fewer model calls alone do not establish better navigation.


\subsection{Quantitative Results}\label{sec:nav-results}

\paragraph{Success across scene conditions}
Table~\ref{tab:main-results} shows that WAVE-Go has the highest reported SR in all three conditions and the highest ID SPL. Relative to the strongest evaluated baseline, the absolute SR gains are 4.7\% in ID, 5.1\% in scene-OOD, and 7.7\% in dynamic-OOD; ID SPL improves by an absolute 5.0\%. The larger gain under dynamic-OOD conditions is consistent with the benefits of adaptive prefix selection and observation-based interruption: the former limits execution length as estimated risk increases, while the latter cancels pending commands when new observations invalidate the plan. Together, these mechanisms reduce continued execution of outdated action sequences as obstacles move or execution conditions change.

\paragraph{Collisions and task completion}
WAVE-Go also reduces the collision rate by 34.1\% relative to the strongest baseline. Higher success alongside fewer collisions is more informative than collision reduction alone, which could result from stopping without completing tasks. Nevertheless, the table gives one distance-normalized collision rate per method, rather than separate rates for each scene condition. Collision incidence, total distance, and deadlock counts are still needed to characterize this trade-off fully.

\begin{figure*}[!t]\centering
\includegraphics[width=0.60\textwidth]{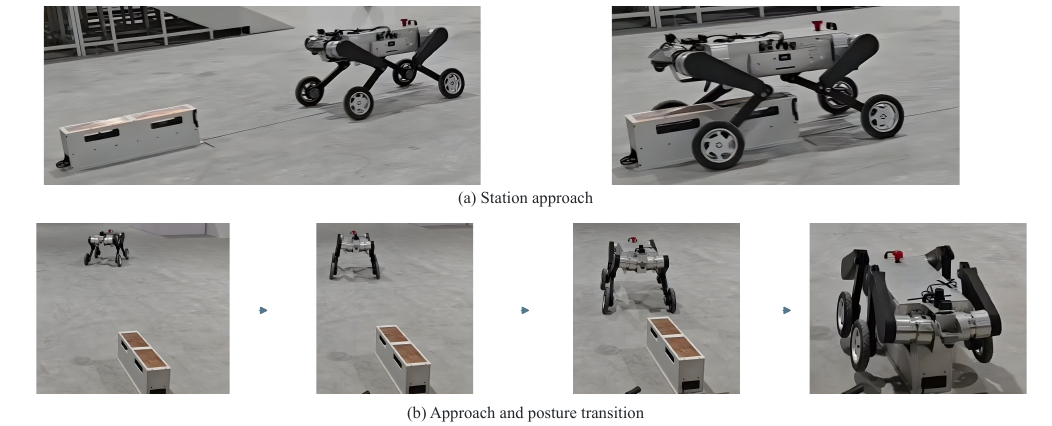}
\caption{Real-world station approach and posture-transition experiment.}
\label{fig:real-charging}
\vspace{-8pt}
\end{figure*}

\begin{figure}[!t]
\centering
\includegraphics[width=\columnwidth]{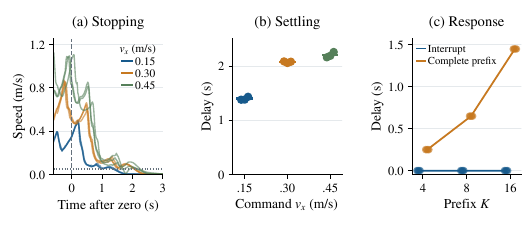}
\caption{LUNA stopping measurements in simulation: (a) speed after zero commands, (b) settling delay at different commanded speeds, and (c) cancellation delay across prefix lengths with and without interruption.}
\label{fig:luna-stopping}
\vspace{4pt}
\end{figure}

\subsection{Execution-Policy Ablations}\label{sec:execution-results}

\paragraph{Fixed execution lengths}
Table~\ref{tab:ablations} exposes the cost of selecting one execution length for all conditions. Increasing $K$ from 4 to 16 reduces replanning from 48.6 to 17.8 invocations per 100\,m and lowers command variation, but decreases success from 70.1\% to 66.0\% and increases collisions from 3.1 to 6.7 per 100\,m. Thus, fewer planning calls and smoother commands alone do not imply better navigation. Reaction latency also increases with $K$, although all three variants retain runtime interruption. Explaining this trend requires examining the timing of observations and command dispatches.

\paragraph{Adaptive selection}
Table~\ref{tab:ablations} shows that adaptive selection improves navigation outcomes while reducing replanning frequency compared with short fixed-prefix execution. Relative to fixed $K=4$, WAVE-Go increases success by an absolute 4.0\% and reduces replanning by 51.2\%, with fewer collisions per 100\,m (2.9 versus 3.1). The trade-off is a 0.03\,s increase in reaction p95, from 0.18 to 0.21\,s. Thus, WAVE-Go more than halves planning calls per unit distance while improving task completion, with a modest increase in command-response latency.

WAVE-Go also outperforms both single-cue adaptive selectors across all metrics in the table. Compared with world-uncertainty-only selection, it improves success by an absolute 2.2\%, reduces collisions by 14.7\%, and lowers both replanning frequency and reaction p95. This pattern is consistent with the benefit of combining predictive cues with execution conditions: uncertainty alone does not fully describe whether a command remains executable.

\begin{table}[!t]
\centering
\caption{Additional measurements of action selection.}
\label{tab:real}
\footnotesize
\setlength{\tabcolsep}{3pt}
\begin{tabular*}{\columnwidth}{@{\extracolsep{\fill}}ccc@{}}
\toprule
\shortstack{Episode\\duration (s)} & \shortstack{Action-selection\\rounds} & \shortstack{Mean selection\\rate (Hz)}\\
\midrule
297.891 & 121 & 0.4062\\
\bottomrule
\end{tabular*}
\end{table}

\paragraph{Runtime interruption}
Reaction latency measures the time to issue a stop or replacement command. Comparing full WAVE-Go with the no-interruption variant evaluates the effect of cancelling pending commands when execution conditions change. Enabling interruption improves success by an absolute 3.6\%, reduces collisions per 100\,m by 45.3\%, and lowers reaction p95 by 61.1\%, at the cost of 17.9\% more replanning calls per 100\,m. By allowing new observations to trigger cancellation and replanning, runtime interruption improves task completion and responsiveness while reducing collisions, with a corresponding increase in planning frequency.

\subsection{Qualitative Results}\label{sec:qualitative}

Figure~\ref{fig:sim-exploration} presents simulation experiments across robot embodiments and navigation settings. Panels (a) and (b) show Go2-W household search and LUNA corridor navigation, respectively. Panel (c) shows a continuous LUNA physics rollout through red pillars in MuJoCo, driven by a waypoint controller and the official walking policy. Panel (d) illustrates Go2-W stair and obstacle traversal. Navigating narrow passages requires checking whether upcoming commands remain feasible as the robot approaches obstacles, while traversing stairs and uneven terrain requires accounting for changes in posture and locomotion mode. These requirements motivate command checks based on updated observations and mode-dependent execution constraints in WAVE-Go.

Figure~\ref{fig:real-charging} shows a real-world experiment in which the robot approaches the station, aligns with it, and lowers its posture. The experiment illustrates WAVE-Go's ability to transition from navigation to a task stage requiring a specific robot posture or configuration.

\subsection{Additional Measurements}\label{sec:prototype}

Table~\ref{tab:real} summarizes action-selection measurements from a long-horizon navigation experiment on Go2-W using a Cosmos3-Edge model \cite{nvidia2026cosmos3} on an NVIDIA L20s GPU. The episode lasted 297.891\,s over 121 action-selection rounds, with a mean selection rate of 0.4062\,Hz. Figure~\ref{fig:luna-stopping} presents separate LUNA simulation trials, distinct from the Go2-W episode in Table~\ref{tab:real}, to distinguish command cancellation from subsequent body settling. Across nine stopping trials, settling takes 1.38--2.26\,s after zero-command publication, with longer delays at higher forward-command levels. In 18 scheduled-invalidation probes, waiting for prefix completion yields mean cancellation delays of 0.25, 0.65, and 1.45\,s for $K=4,8,16$, respectively; interruption removes this wait while independent emergency checks remain active. These probes measure local software response to a scheduled event, excluding sensing and physical braking. These measurements support WAVE-Go's interruptible execution design by showing that pending commands can be cancelled without waiting for the selected prefix to finish. Notably, command cancellation remains prompt even with longer execution prefixes. For more details, see the supplementary videos.

\FloatBarrier
\section{Conclusion}

We presented WAVE-Go, an image-goal navigation framework that connects world--action prediction to adaptive, interruptible execution for wheel-legged robots. Its executor selects action prefixes under an estimated cumulative failure budget and command-validity constraints, while updated observations can cancel pending commands. Evidence-gated posture and locomotion-mode transitions require confirmation of the resulting configuration and renewed prediction before subsequent motion. This design makes execution length a feedback-informed decision, allowing the robot to adjust its commitment to a predicted action sequence as conditions change.

The navigation results show higher success across the evaluated scene conditions and fewer collisions, including an absolute 7.7\% gain in dynamic-OOD success and a 34.1\% reduction in collisions per 100\,m over the strongest baseline, together with an absolute 4.0\% success gain and 51.2\% fewer replanning calls relative to short fixed-horizon execution. Ablations further show that runtime interruption improves success, collision rate, and reaction latency at the cost of additional replanning. These results highlight the value of adapting execution commitments to changing conditions while retaining feedback throughout an action sequence.

\FloatBarrier
\bibliographystyle{IEEEtran}
\bibliography{wave_go_v3}

\end{document}